%% file: main-nocomments.tex
\documentclass[runningheads]{llncs}

\usepackage[T1]{fontenc}

\usepackage{graphicx}

\usepackage{hyperref}
\usepackage{cleveref}
\usepackage{booktabs}

\usepackage{color}

\usepackage[normalem]{ulem}  
\usepackage{xcolor}

\def\camera#1{} 

\begin{document}

\title{
Corpus of Cross-lingual Dialogues with Minutes and Detection of Misunderstandings}

\titlerunning{Corpus of Cross-lingual Dialogues}

\author{Marko Čechovič\orcidID{0009-0006-2549-7545} \and
Natália Komorníková\orcidID{0009-0000-8061-3103} \and \\
Dominik Macháček\orcidID{0000-0002-5530-1615} \and
Ondrej Bojar\orcidID{0000-0002-0606-0050}
}

\authorrunning{Čechovič et al.}

\institute{
Charles University, Faculty of Mathematics and Physics, \\ Institute of Formal and Applied Linguistics (ÚFAL),
Prague, Czechia
\email{marko.cechovic@matfyz.cuni.cz},
\email{\{machacek,bojar\}@ufal.mff.cuni.cz}}

\maketitle

\def\TotalLangs{12}

\begin{abstract}

Speech processing and translation technology have the potential to facilitate meetings of individuals who do not share any common language. To evaluate automatic systems for such a task, a versatile and realistic evaluation corpus is needed. Therefore, we create and present a corpus of cross-lingual dialogues between individuals without a common language who were facilitated by automatic simultaneous speech translation. The corpus consists of 5 hours of speech recordings with ASR and gold transcripts in \TotalLangs{} original languages and automatic and corrected translations into English. For the purposes of research into cross-lingual summarization, our corpus also includes written summaries (minutes) of the meetings.

Moreover, we propose automatic detection of misunderstandings. For an overview of this task and its complexity, we attempt to quantify misunderstandings in cross-lingual meetings. We annotate misunderstandings manually and also test the ability of current large language models to detect them automatically. The results show that the Gemini model is able to identify text spans with misunderstandings with recall of 77\% and precision of 47\%.

\end{abstract}

\def\InCroMin{InCroMin}   
\section{Introduction}

In today's globalized world, there is a growing need for effective interaction across languages. Meetings and discussions between individuals of diverse linguistic backgrounds are becoming more frequent. Speech processing and translation technology have the potential to facilitate these interactions and enable each person to communicate in the language of their preference. 

In recent years, significant advances have been made in Automatic Speech Recognition (ASR), Speech Translation (ST), Simultaneous Speech Translation (SST), and meeting summarization or ``minuting.''
However, to ensure that automatic systems can accurately support the cross-lingual interactions, a realistic and versatile evaluation corpus is needed. Current evaluation frameworks often rely on simplified setups, such as read speech instead of spontaneous \cite{Panayotov2015LibrispeechAA}, or processing isolated sentences retrieved from monologues \cite{must-c}. The complex real-life challenges of the cross-lingual dialogues such as spontaneity, interactivity, specific terminology, background knowledge of communication partners, 
the missing sentence segmentation, inter-sentence context, sentences completed by another party in the meeting, and recording conditions matching the frequent real-life use cases are often not covered by the existing datasets.

To address this gap, we create and present \InCroMin, a unique corpus of cross-lingual dialogues between individuals who do not share any common language. These dialogues were facilitated by a state-of-the-art automatic simultaneous speech translation system. This corpus enables evaluation of various aspects of spoken language processing, including ASR, ST, SST, Quality Estimation, and Automatic Minuting in \TotalLangs{} languages other than English. We present \InCroMin{} release with 5 hours of audio of 24 participants.\footnote{\InCroMin{} corpus is available at \url{http://hdl.handle.net/11234/1-5956}.}

Moreover, effective communication is not only concerned with conveying information, but also with ensuring that the other parties understand. Misunderstandings can arise due to various factors, including technical issues, automatic processing errors, language ambiguity, the differences in culture and personal backgrounds, etc. In this paper, we also initiate the study of misunderstandings in such a cross-lingual setting by annotating misunderstandings in \InCroMin{} meetings and proposing an automatic detection using the current Large Language Models (LLMs). Such an automatic system monitoring the interaction could alert when there is a risk of misunderstanding and propose mitigation actions, such as clarification questions. Our initial experiment shows that the LLM Gemini\footnote{\url{https://gemini.google.com/}} is able to detect misunderstandings in English text translations of dialogues with recall of 77\% and precision of 47\%. This result is promising for further work in this area.
T

The structure of this paper is as follows: In \Cref{sec:corpus}, we present how we create the \InCroMin{} corpus of cross-lingual dialogues with volunteers who had calls using a speech translation tool Minuteman, what data we collected and how we processed them. Then, in \Cref{sec:mu}, we present the annotation of misunderstandings and the detection by a Large Language Model. In \Cref{sec:feedback}, we present users' feedback on Minuteman. We summarize the problems of a current state-of-the-art system for simultaneous translation of cross-lingual dialogues. We conclude in \Cref{sec:conclusion}.

\section{\InCroMin{} Corpus}
\label{sec:corpus}

\def\ul#1{\underline{#1}}
\def\furl#1{\footnote{\url{#1}}}

We name the corpus \InCroMin{}, which is an acronym for \ul{In}cremental \ul{Cro}ss-Lingual \ul{Min}utes. As ``minutes'' we mean a brief written record that summarizes the important points of a meeting, such as the agenda, conclusions, and action items \cite{TSDghosal-etal-2023-overview}. ``Cross-lingual'' means that the meeting participants use different languages and do not understand each other directly. The meeetings are facilitated by a speech processing and translation system that translates the participants' speech and written minutes into the participants' languages of preference. Incremental means that the translations and the minutes grow incrementally, simultaneously with the meeting. 

In this section, we describe the collection of the \InCroMin{} corpus: first, in \Cref{sec:technical}, the necessary technical background -- the Minuteman tool that facilitates cross-lingual meetings -- and then data creation and collection (\Cref{sec:data}), post-processing (\Cref{sec:post}), and corpus overview (\Cref{sec:over}). 

\subsection{Technical Setup}
\label{sec:technical}

The cross-lingual meetings in \InCroMin{} were held online using the Fairmeeting\furl{https://fairmeeting.net/} videoconferencing platform and Minuteman \cite{Minuteman} tool for simultaneous translation, minuting, and data collection.

Minuteman is able to collect speech audio from the Fairmeeting session in individual audio tracks, one for each participant in the meeting. Therefore, speaker identification is implied and no automatic diarization is needed. Minuteman applies Whisper \cite{Whisper} Large-v3 speech translation model for simultaneous translation from the original language into English. 

For \InCroMin{}, we extended Minuteman with the NLLB machine translation model \cite{NLLB},\footnote{Version: facebook/NLLB-200-distilled-600M.} for multi-target translation from English to the languages selected at the meeting, and with data collection support.
\camera{\footnote{The extended Minuteman version is available at \url{https://github.com/fkmjec/minuteman}.}}

\def\bold#1{\textbf{#1}}

\subsection{Data Creation}
\label{sec:data}

For the \InCroMin{} corpus, we organized meetings of 2-5 participants. All participants were volunteers selected primarily from the personal and professional network of the authors. Due to the limited time and resources for this study, we did not balance the language distribution of the participants, although it coincidently led to a higher representation of Czech speakers.

We matched the participants together in a way that (1) at the meeting, there were at least two groups of participants that used different languages that were not known to the other group, and (2) they had a common topic to discuss, for example, consultation of a student project, discussing common topics, a casual chat between friends, etc.

We asked participants not to use their common language, although many of them were able to speak English to each other. 
The cross-linguality in the meetings is therefore not a genuine need; however, genuine dialogues between people who do not understand each other's language are currently very rare and hard to collect. Therefore, we consider our arrangement as the most feasible way to collect the cross-lingual dialogues.

All meetings are authentic conversations between real people, except for avoiding the common language and except for two simulated meetings. They simulate an interview between a refugee and an integration center officer. These simulated meetings were run by two integration center officers, so they closely resemble real meetings of this kind. Simulating such dialogues is a feasible way to collect data from the refugee support domain because asking authentic refugees i
for data collection consent is usually not feasible.

The languages used by the participants during the meeting were selected according to their preference from the list of supported languages, which included 200 languages supported by the NLLB machine translation model. We asked participants to select and use one primary language during the meeting. In total, \TotalLangs{} languages were used. 
No participant selected English as the primary language because, coincidently, there was no other participant who would not understand English.
n

Participants were informed of the collection and planned release of their data
after removing personal information such as names, but including their voice.
Since voice is a personal identifiable according to the national regulation, the
participants had to provide their informed consent with regard to processing and release of data.

Participants were asked to hold the meeting for 30 minutes. After the meeting, they were asked to complete a feedback form in which they reported their subjective evaluation of the technical setup. The questionnaire took around 15 minutes. Some participants were involved in multiple meetings.

\subsection{Post-processing}
\label{sec:post}

We post-processed the meeting data with the following steps:

\begin{enumerate}
    \item Retrieval: We retrieved data from Minuteman, specifically the audio tracks of each participant and the automatic translations in their language that were displayed to them during the meeting. They may be used in further analysis.
    \item Synchronization and trimming: we synchronized the audio tracks of the speakers at the meeting and trimmed them so that they contain only the part that forms the meeting, excluding e.g.\ waiting for the other participant or setting up Minuteman. We also trimmed the log of displayed translations accordingly.
    \item Automatic transcriptions and translations: Then we processed the audio with ASR and translated the speech into English. We report which automatic system was used for further analysis.
    \item Corrections: The ASR transcripts and automatic speech translations were voluntarily corrected by the participants or by a paid annotator who is fluent in the language.
    \item Deidentification: Information such as names, locations, and organizations that could identify the private persons in the meeting (the speakers and their peers, but not e.g.\ authors of publications) are removed from the texts and replaced with a placeholder. We remove the audio segments where this information was disclosed by replacing them with silence and reporting the time intervals. This step was processed by an annotator who is proficient in the spoken language.
    \item Minutes: An annotator created minutes for the meeting in English.
    \item Misunderstandings annotation. See \Cref{sec:mu}.
\end{enumerate}

\subsection{Corpus Overview}
\label{sec:over}

\Cref{tab:overview} summarizes the total corpus size. At the time of publication, we release 10 meetings as the ``first'' release. There are 24 participants (one meeting is with 5 participants, one with 3, the others with 2) covering 12 languages. In total, there are 296 minutes of audio. Some of the corrected transcripts and translations are pending. There are 15 corrected transcripts of 24 (63\%) and 13 translations into English (54\%). The remaining corrections will be added in future releases.

Additionally, we plan to release another 10 meetings with 20 participants using 4 more languages that are not in the first release. These data will be released as soon as the deidentification is complete.

\begin{table}[]
    \centering
    \caption{Size statistics of the first \InCroMin{} release and of the planned release. The duration is in minutes and includes silence.}
    \begin{tabular}{l|ccccc|cccc}
&&&&&& \multicolumn{2}{c}{Corrected} \\
Release & Meetings & Participants & Languages & Duration & Words & Transcripts & Translations \\
\hline
\textbf{First} & \textbf{10} & \textbf{24} & \textbf{12} & \textbf{296m} &  \textbf{26\,854} & 15 (63\%) & 13 (54\%) \\
Planned & 10 & 20 & +4 & 300m &          &  6   & 7            \\
\hline
Total & 20 & 44 & 16 & 596m & & 21 & 20 \\
    \end{tabular}
    \label{tab:overview}
\end{table}

Since language distribution is an important aspect of multilingual corpora, we ask how much time was each language spoken in the corpus, and how many words were uttered in each language. For that, we first processed all audios with Silero Voice Activity Detection \cite{Silero-VAD}, which marked voiced segments for each participant interleaved with silence, especially when both parties were waiting for simultaneous translation to appear. For estimating the language distribution, we split the silent intervals into halves and assign them to the language adjacent to them. 

To count the words in each language, we processed the best available transcripts, which is the corrected one if available, or ASR if not, with SacreMoses tokenizer. It applies segmentation on spaces using rules for space-delimiting languages, and morphological segmentation for languages that do not use spaces, such as Chinese. We report the number of tokens. Note that the number of tokens is affected by the quality of ASR, which is lower for some languages. 

\Cref{tab:lang-distr} summarizes the distribution of languages in the released part of \InCroMin{}. Czech represents 45\% of the corpus (133 minutes of 296) because we found more Czech-speaking volunteers than others. Nine languages are represented by at least one participant in one two-person meeting (11 to 34 minutes). Two languages are represented by 5 and 3 minutes by participants in the meeting with 5 participants. These two participants spoke less than the others in the 30-minute meeting. 

\begin{table}[]
    \centering
\caption{Distribution of languages in \InCroMin{} as the number of transcript tokens (corrected if available, ASR otherwise) and number of participants in meetings using the language primarily.}
    \begin{tabular}{l|rr| r}
\textbf{Language} & \textbf{Duration} & \textbf{Tokens} & \textbf{Participants} \\
\hline
Czech & 133m & 12\,213 & 10\\
Russian & 34m & 3\,356 & 2\\
Chinese & 20m & 2\,816 & 2\\
Brasilian Portuguese & 18m & 1\,698 & 1\\
Slovak & 14m & 1\,206 & 2\\
French & 15m & 1\,195 & 1\\
Italian & 15m & 1\,038 & 1\\
Ukrainian & 14m & 936 & 1\\
Vietnamese & 11m & 927 & 1\\
Marathi & 15m & 779 & 1\\
Portuguese &  5m & 452 & 1\\
Spanish & 3m & 238 & 1\\
\hline
Total &  296m & 26\,854 & 24 \\
    \end{tabular}
    \label{tab:lang-distr}
\end{table}

\section{Detection of Misunderstandings}
\label{sec:mu}

As an initial study of automatic detection of misunderstandings in the cross-lingual call, we first propose a protocol to annotate the misunderstandings (\Cref{sec:annotation-of-misunderstandings}). Second, present the results of annotations (\Cref{sec:ann-results}), and measure the inter-annotator agreement of two human annotators (\Cref{sec:iaa}). Then, in \Cref{sec:gemini-capabilities}, we apply a state-of-the-art generative large language model Gemini to detect misunderstandings, evaluate it, and discuss the results.

\subsection{Annotation Protocol}
\label{sec:annotation-of-misunderstandings}

The annotation protocol defines instructions for human annotators to detect and classify misunderstandings in the text record of all parties in the meeting. Since there are many languages in the corpus and the annotators are not proficient in so many languages, they were processing the revised English translations of the dialogues. T

The annotation process consists of the following steps:

\begin{enumerate}
    \item \textbf{Identification of markables:} The annotator searches for a segment of the dialogue where a misunderstanding has occured. This is called a ``markable.''
    \item \textbf{Delimiting markable spans:} The annotator marks the lines of text (usually several sentences), called ``spans,'' where misunderstandings occured.
    \item \textbf{Awareness of misunderstading:} The annotator marks each misunderstanding as ``acknowledged'' or ``unrecognized,'' depending whether the speakers in the dialogues reacted in a way that seems that they were aware of the misunderstanding shortly after it occurred. For example, whether they repeated what was said, or asked for clarification.
    \item \textbf{Reason of misunderstanding:}
For each markable, an annotator classifies the reason for misunderstanding. Each reason was attributed to the speaker A or B.
\begin{itemize}
     
    \item \textbf{Translation error:}  
    The translation between the A and B, or the ASR transcript of A or B, was problematic. For example, the receiving speaker comments on garbled or confusing output.
    \item \textbf{Delay:} The simultaneous translation and transcription has a processing delay, which sometimes caused confusion. For example, speakers repeated their last utterance because they assumed that it was not processed at all or reacted to the last displayed utterance that was not the last that the other speaker produced.
    \item \textbf{Technical problem:} Issues such as Internet connectivity or Minuteman server errors caused communication problems.
    \item \textbf{Genuine misunderstanding:} Misunderstanding from other reasons than technological or processing errors. This label is assigned to reasons that we presume would also occur in a monolingual dialogue. For example, referring to an entity that is not familiar enough to the other party.
    
\end{itemize}
      \item \textbf{Gemini detection and revision:} The records are processed by Gemini LLM that is prompted to automatically detect the misunderstandings markables. If it is found by Gemini, the annotator marks it as ``true positive'' or ``false positive,'' depending whether the annotator considers it a misunderstanding or not.  
\end{enumerate}

\subsection{Annotation Results}
\label{sec:ann-results}

The data that were annotated for misunderstandings are the revised English translations of 14 meetings from both the first and planned \InCroMin{} release. All the meetings were of two participants. 

\Cref{tab:annotations} contains a summary of annotations. Across the 14 selected meetings, we acquired 25 annotations, where the two annotators together found 222 misunderstandings. Most of them, 36.5\%, were attributed to translation errors, 29.7\% to genuine misunderstandings, 14.4\% to system delays, and 19.4\% to technical errors. The proportion of acknowledged misunderstandings was 57\%.

Since the quality of speech and translation processing is very dependent on the source and target languages, we observe the difference in proportions of reasons and awareness of misunderstandings in the 9 meetings where Czech was spoken vs.\ in the 5 meetings without any Czech speaker. In the Czech meetings, there were 8\% fewer misunderstandings attributed to translation errors, 8\% fewer genuine misunderstandings, and 14\%  more misunderstandings due to technical errors. We hypothesize that many Czech meetings were held by a professor and his students who had large shared background knowledge, and thus a better mutual understanding. For the same reason, they could also be 17\% more aware of misunderstandings.

\begin{table}[]
\centering
\caption{Sum of misunderstandings (MU) annotations of two annotators on all 14 meetings (first row) and in the meetings where Czech was spoken by one participants vs.\ meetings without Czech.}
\begin{tabular}{lr|rr|rrrr|rr}
 
& & & \textbf{} &  \multicolumn{4}{c|}{\textbf{Reason \%}} & \multicolumn{2}{c}{\textbf{Awareness \%}} \\
 
\textbf{Meetings} & \textbf{num.} & \textbf{Ann.} 
& \textbf{MU} & Gen. & Transl. & Delay & Tech. & Ack. & Unrec. \\
\hline
All & 14 
& 25 & 222 & 29.7 & 36.5 & 14.4 & 19.4 & 57.1 & 42.9 \\
\hline
with Czech & 9 & 15
& 138 & 26.8 & 33.3 & 15.2 & 24.6 & 64.0 & 36.0 \\
without Czech & 5  
& 10 
&  84 & 34.5 & 41.7 & 13.1 & 10.7 & 47.1 & 52.9 \\
\end{tabular}
\label{tab:annotations}
\end{table}

\subsection{Inter-Annotator Agreement}
\label{sec:iaa}

Two independent annotators annotated 11 of the 14 meetings. We make use of
this double annotation to assess the inter-annotator agreement (IAA).
Due to the small size of the dataset and the nature of the annotation (each annotator is identifying their own set
of misunderstandings), we consider IAA only at the level of overall counts, not at the
level of individual misunderstanding items. In other words, we count how often
an annotator reported e.g. a technical problem in a given meeting, and compare
this number with the count of technical problems reported by the other
annotator. For a given meeting and annotation category, we plot the count
reported by one annotator on the x-axis and the
count reported by the other annotator on the y-axis. We order the pair of
annotators so that the x-value is always the lower one in the pair.

\begin{table}[]
    \centering
        \caption{Pearson correlation coefficients for counts of misunderstanding items reported for the given misunderstanding type (upper part) or reason (lower part). ,
        }
\begin{tabular}{l|r}
\bf Misunderstanding Type or Reason & \bf Pearson Coefficient \\
\hline
\input{iaa-pearson.tex}\end{tabular}
    \label{tab:iaa-pearons}
\end{table}

For each of the observed quantities, we also calculate the Pearson correlation coefficient, see \Cref{tab:iaa-pearons}.
We see that in many categories, the correlations are reasonably good, e.g. 0.94 for attributing the misunderstanding to a translation error or 0.72 for spotting a technical problem.
The annotators seem to disagree when annotating Gemini false positives (FP),
i.e.\ items that Gemini identified as misunderstandings but one of the annotators
does not think so. The low correlation can be traced to 5 meetings 

in each of which one annotator disapproved 4 misunderstandings suggested by Gemini
but the other annotator did not reject any Gemini suggestions.

\subsection{Gemini Capabilities in Identifying Misunderstandings}
\label{sec:gemini-capabilities}

Given the recent good abilities of LLMs in general text processing and assessment tasks, we wanted to evaluate if LLMs would be reliable in our annotation of misunderstandings.

We ran Gemini 1.5 Pro LLM from Google in our own private Google Cloud environment to ensure privacy of our data. We could have chosen other models such as GPT-3.5 turbo or Llama 3 but since we experienced difficulties in running those models in a privacy preserving manner and since Gemini 1.5 Pro has the longest context of them all, we decided to proceed with Gemini.

After experimenting with several variants of prompts, we proposed a simple and effective prompt for the language model, the one in \Cref{fig:gemini-prompt}. Then, we appended the meeting conversations with timestamps, one meeting at a time, and processed it with the model. The model was
able to identify most of the misunderstandings with precise timestamps and also
provide reasoning.

To evaluate the quality of Gemini annotations, we provided our annotators with the outputs of Gemini 1.5 and asked them to validate Gemini annotations \emph{after} they had done their own annotation.

\begin{figure}[]
    \centering
        \caption{Gemini 1.5 Pro system prompt for finding misunderstandings.}
\fbox{
\begin{minipage}{.9\textwidth}
\small
You are a experienced language and conversation expert.
The user will give you an annotated transcription of a conversation between two speakers. Identify sections of the transcript where some misunderstanding happened.
For example where someone needed to ask if the other speaker understood or if one speaker was talking about something and the second one was answering/continuing the conversation but the topics did not match at all. Create summary of the findings and put example with timestamps to each misunderstanding found.
\end{minipage}
}
    \label{fig:gemini-prompt}
\end{figure}

\begin{table}[]
\centering
\caption{Confusion matrix of validity of Gemini's classification of misunderstandings.
}
\begin{tabular}{ccc} \toprule & \multicolumn{2}{c}{Gemini 1.5 Pro} 
\\ \cmidrule{2-3} Annotators & True & False\\ \midrule 
True & 122 & 137\\ 
False & 36& -\\ 
\bottomrule 
\end{tabular}
\label{tab:table-Gemini}
\end{table}

\Cref{tab:table-Gemini} provides the confusion matrix between Gemini and annotators identification of points of misunderstandings.
Based on these results, we evaluate how Gemini performs in a task of finding misunderstandings in a dialogue. We  estimate precision 47\% and recall 77\%.

Although we used all available data in this evaluation in this initial experiment, we rather warn against generalization of Gemini's performance only from this small dataset.  
We also highlight that our method is not able to find false positives. 
Searching for false positives is not feasible because Gemini usually detects one misunderstanding in a large text chunk, where the annotators find several misunderstandings. This was one of the biggest problems with using Gemini for misunderstandings detection.

\section{Users' Feedback on the Dialogue Translation Tool}
\label{sec:feedback}

Collection of \InCroMin{} corpus was an opportunity to collect feedback from prototypical users on the state-of-the-art technology for facilitating cross-lingual meetings. Therefore, after each meeting, the participants were asked to report on their personal experience with Minuteman, their expectations and quality of service in a questionnaire.

In this section, we summarize and highlight the key observations we received.

\Cref{tab:langs-and-usability} shows the assessment of usability of the tool. We see that very few (2 in fact) respondents are pessimistic about the tool usability. 44\% of the respondents suggested that some fixes are needed and more than a half of the participants considers the system usable at the current stage of development.

\begin{table}[]
    \centering
    \caption{Users' assessment of overall usability of Minuteman as a tool for facilitating cross-lingual dialogues. 45 answers in total.}
    \begin{tabular}{r|p{0.8\textwidth}}
\textbf{Answers \%} & \textbf{Assessment} \\
\hline
15.6\% & Yes, it was usable already now. \\
\hline
35.6\% & It was usable but tedious. \\
\hline
44.4\% & If the following things were fixed, it would be usable. \\
\hline
4.4\% & I don't think any too can fix this, I would rather avoid such a multilingual meeting than to use this tool.  
    \end{tabular}
    \label{tab:langs-and-usability}
\end{table}

\begin{figure}[]
    \centering
    \caption{Subjectively assessed slowdown due to cross-lingual barrier (left) and level of tiredness compared to normal calls featuring one language (right).}
    \includegraphics[width=0.49\linewidth]{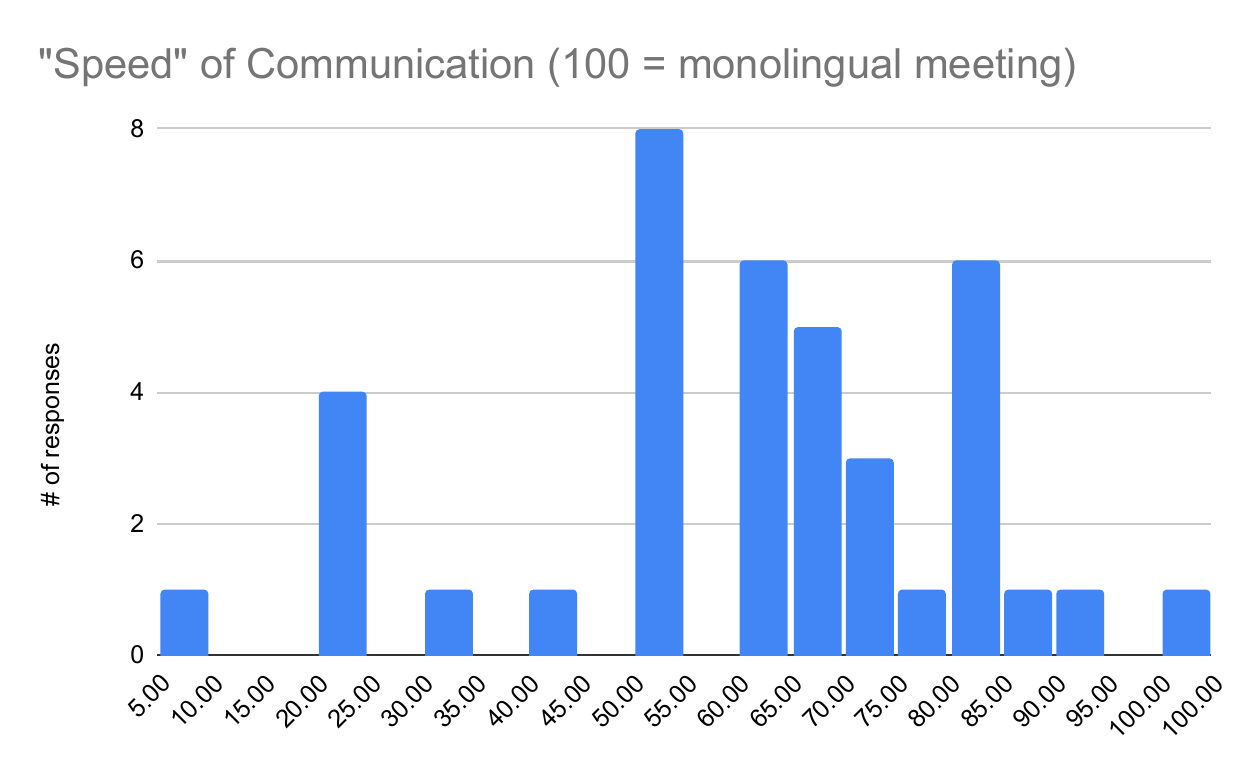}
    \hfill
    \includegraphics[width=0.49\linewidth]{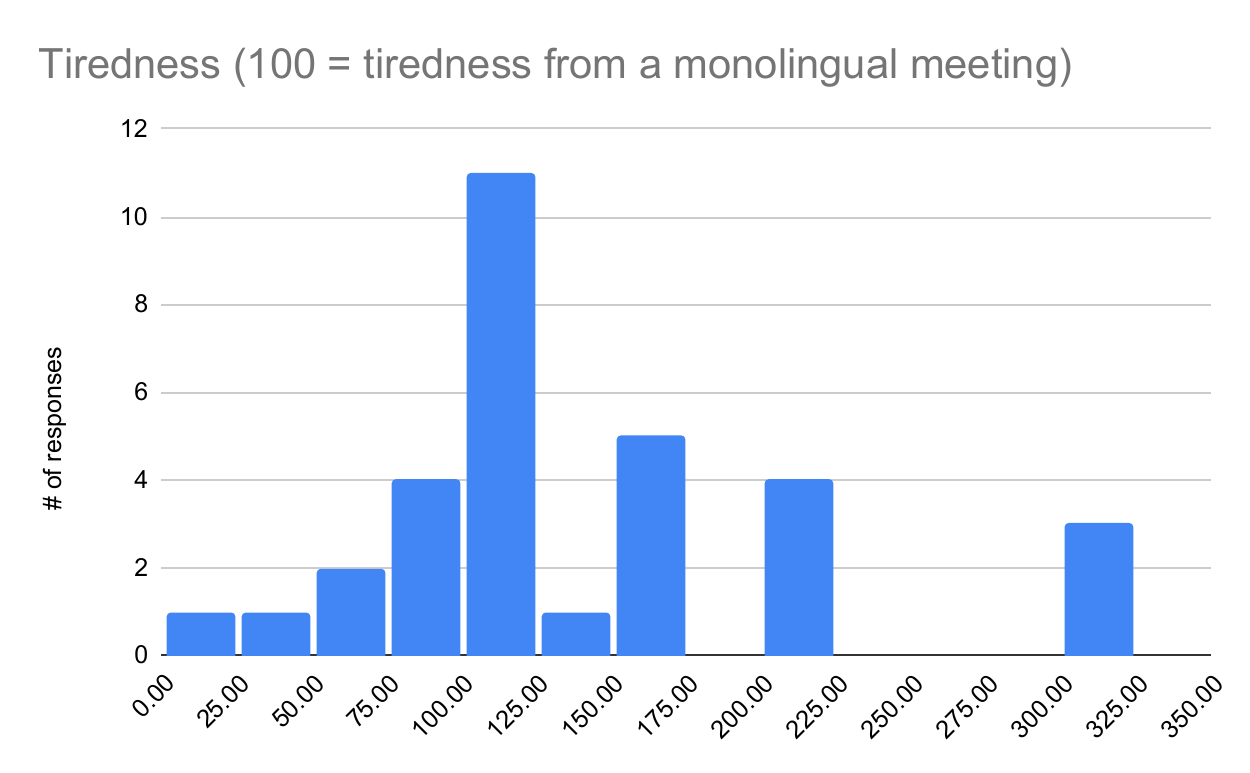} 
    \label{fig:speed-and-tiredness}
\end{figure}

Separately, we asked the call participants to assess a ``slowdown'' caused by the language barrier in the call, and the subjective level of tiredness relative to a single-language call, see \Cref{fig:speed-and-tiredness}. We see that most participants do realize a considerable slowdown, typically around the factor of 2, although there are also many calls reaching 70--80\% of normal calls ``speed''. This ``speed'' was only self-assessed, essentially based on the plan for the call that the participants should have made, and the actual ability to get their messages across to the other speaker. A more controlled setting would be however needed for a reliable result in this respect.

The ``tiredness'' was also only subjectively assessed and most participants do not feel any extra burden, although there are striking exceptions like 3$\times$ as tired. For a reliable interpretation of this result, a contrastive test would be however needed, featuring meetings where there is no language barrier but the modality matches our setting: speaking and only reading, not listening to the other participants.

The feedback that the users delivered in text form highlights speech processing and translation problems such as wrong target language of machine translation, even when it was set correctly, and errors caused by audio segmentation and voice activity detection.

Some users commented on serious problems with wrong translation of ``dialogue management language'' or ``meta-language.'' It is difficult to lead a dialogue when the system wrongly translates the question who is going to speak first or similar, or even such a translation error seemingly introduces a new topic. Some users reported problems with incorrect preservation of questions and affirmative sentences, and some reported short affirmative sentences on any sound, which was rather confusing.

Finally, they reported problems with output presentation, such as that it was unclear whether the machine has processed all the speech or more is yet to come, which made turn-taking unclear. Clearer speaker separation in the output was also missing for some users, and some suggested improving synchronization by delaying the other party's audio to make it match the translation as it is appearing. Other comments were on the user interface, such as annoying buttons, text scrolling, or a demand for a push-to-talk button.

\section{Conclusion}
\label{sec:conclusion}

We presented our contribution towards an effective and high-quality tool to facilitate cross-lingual dialogues between people who do not share any common language. We introduced our multi-lingual dialogue corpus \InCroMin{}, which consists of 5 hours of 10 meetings of 24 people speaking 12 languages, as a resource for evaluation of Automatic Speech Recognition, Simultaneous Speech Translation, and Quality Estimation. The minutes included in the corpus may be used for evaluation of automatic meeting summarization tools.

Then, we discussed an initial work towards automatic detection of misunderstandings in the cross-lingual meetings. We show that more than 40\% of misunderstandings are unrecognized by the parties at the meetings in \InCroMin{}, which is an important problem to focus on in further work. We demonstrate the ability of the large language model Gemini to detect misunderstandings in English text transcripts with 47\% precision and 77\% recall. Last but not least, we presented users' feedback on current state-of-the-art simultaneous speech translation for cross-lingual dialogues, which proposes future work, such as focus on the dialogue management language, developing a better user interface, and higher quality speech processing and translation models.

\subsubsection{\ackname}
This work has been supported by Charles University Research Centre program No.\ 24/SSH/009, 
and by Project OP JAK Mezisektorová spolupráce
Nr. CZ.02.01.01/00/23\_020/0008518 named
``Jazykověda, umělá inteligence a jazykové a
řečové technologie: od výzkumu k aplikacím.''
The authors also acknowledge support of
National Recovery Plan project MPO
60273/24\discretionary{/}{}{/}21300/21000 CEDMO 2.0 NPO.

\subsubsection{\discintname}

The authors have no competing interests to declare that are relevant to the content of this article.

\bibliographystyle{splncs04}
\bibliography{utter,incromin-tsd-bibs}

\end{document}

%% file: iaa-pearson.tex
MU acknowledged              &0.83\\
MU unrecognized              &0.90\\
MU found by Gemini           &0.88\\
MU not found by Gemini (FN)  &0.80\\
Gemini TP                    &0.71\\
Gemini FP                    &-0.13\\
\hline
Bad translation              &0.94\\
Delay in ASR                 &0.49\\
Genuine misunderstanding (MU)&0.55\\
Technical problem            &0.72